\pdfoutput=1

\documentclass[11pt]{article}
\usepackage[dvipsnames]{xcolor}

\usepackage{svg}

\usepackage{enumitem}

\usepackage[final]{acl}

\usepackage{times}
\usepackage{latexsym}

\usepackage[T1]{fontenc}

\usepackage[utf8]{inputenc}

\usepackage{microtype}

\usepackage{inconsolata}
\usepackage{soul}

\usepackage{graphicx}

\usepackage{hyperref}       
\usepackage{url}            
\usepackage{booktabs}       
\usepackage{amsfonts}       
\usepackage{nicefrac}       
\usepackage{microtype}      
\usepackage{lineno}         
\usepackage{tcolorbox}
\usepackage{amsmath}
\usepackage{multirow}

\usepackage{enumitem}
\usepackage{xspace}
\usepackage{fdsymbol}

\newcommand{\nolima}{NoLiMa}
\newcommand{\novelqa}{NovelQA}

\title{Tagging-Augmented Generation: Assisting Language Models in Finding Intricate Knowledge In Long Contexts}

\author{
 \textbf{Anwesan Pal\textsuperscript{1}},
 \textbf{Karen Hovsepian\textsuperscript{1}},
 \textbf{Tinghao Guo\textsuperscript{2}},
 \textbf{Mengnan Zhao\textsuperscript{2}},
 \textbf{Somendra Tripathi\textsuperscript{3}},
\\
 \textbf{Nikos Kanakaris\textsuperscript{1}},
 \textbf{George Mihaila\textsuperscript{2}},
 \textbf{Sumit Nigam \textsuperscript{4}}
\\
 \textsuperscript{1}AWS AI Labs,
 \textsuperscript{2}Amazon Web Services,
 \textsuperscript{3}Amazon OTS,
 \textsuperscript{4}Amazon Catalog AI,
\\
 \small{
   \texttt{\{anwesanp, khhovsep, tinghg, mengnanz, somendt, nikosk, georgemh, sumnig\}@amazon.com}
 }
\\
 \small{
 Authors contributed equally
 }
}

\begin{document}
\maketitle
\begin{abstract}
Recent investigations into effective context lengths of modern flagship large language models (LLMs) have revealed major limitations in effective question answering (QA) and reasoning over long and complex contexts for even the largest and most impressive cadre of models.  While approaches like retrieval-augmented generation (RAG) and chunk-based re-ranking attempt to mitigate this issue, they are sensitive to chunking, embedding and retrieval strategies and models, and furthermore, rely on extensive pre-processing, knowledge acquisition and indexing steps. In this paper, we propose Tagging-Augmented Generation (TAG), a lightweight data augmentation strategy that boosts LLM performance in long-context scenarios, without degrading and altering the integrity and composition of retrieved documents.  We validate our hypothesis by augmenting two challenging and directly relevant question-answering benchmarks -- \nolima~and \novelqa~ -- and show that tagging the context or even just adding tag definitions into QA prompts leads to consistent relative performance gains over the baseline -- up to 17\% for 32K token contexts\footnote{As of the time of this publication, our best performing model ranks \#2 in the \nolima~\href{https://github.com/adobe-research/NoLiMa/tree/e84ca973d39ae92e4c521a941fba99fb5ebe56d5?tab=readme-ov-file\#results}{leaderboard} for 32K context.}, and 2.9\% in complex reasoning question-answering for multi-hop queries requiring knowledge across a wide span of text. Additional details are available at \url{https://sites.google.com/view/tag-emnlp}.

\end{abstract}

\begin{figure}[t]
    \centering
    \includegraphics[width=0.85\linewidth]{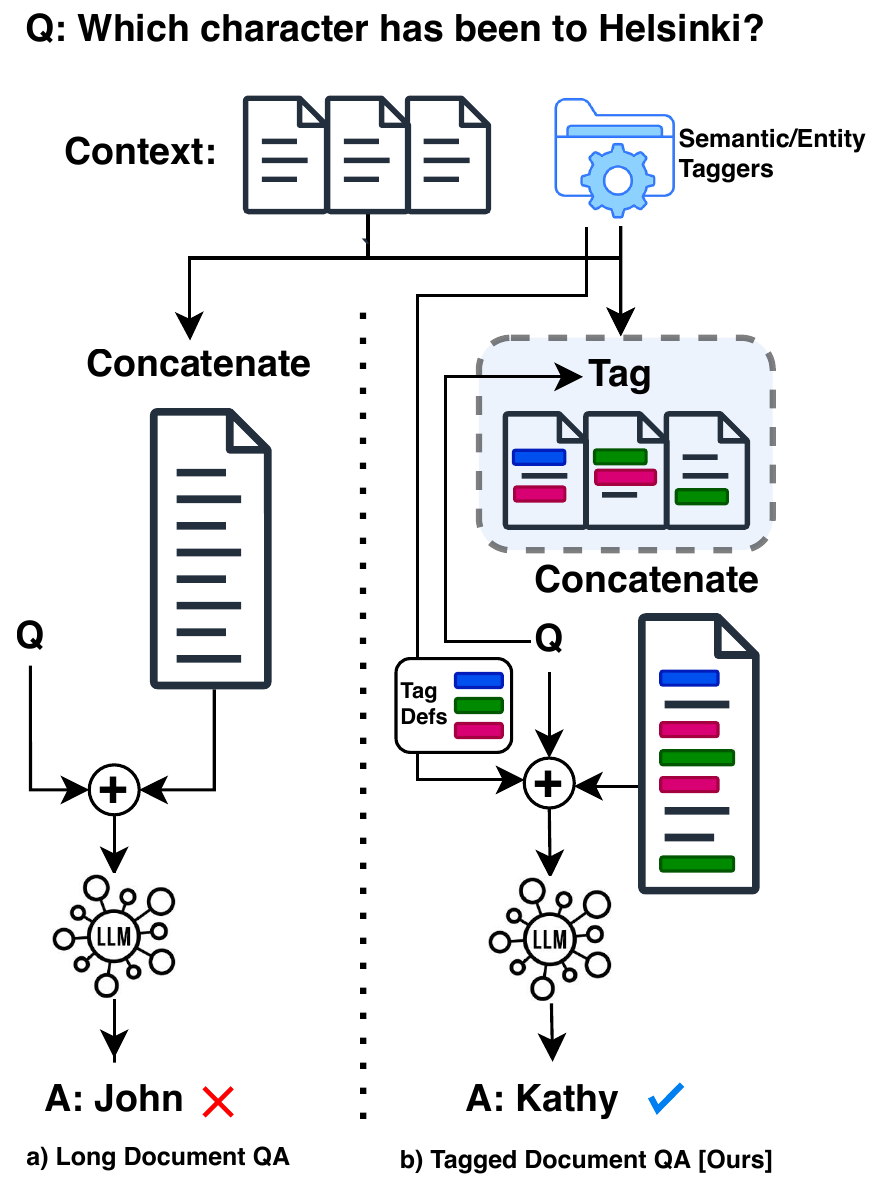}
    \caption{High-level diagram comparing key steps in \textbf{(a)} traditional long-document QA and \textbf{(b)} the proposed tagging-augmented QA. The diagram highlights the plug-and-play use of imported, domain-specific taggers in an agentic flow. Our framework involves semantically enriching context by means of tagging. Furthermore, we also augment the prompt with an imported list of semantic and entity tag names and definitions to guide the LLM towards recognizing such semantic entities.}
    \label{fig:tagthenanswer}
\end{figure}

\section{Introduction}
Latest advances in large language models (LLMs) have dramatically expanded their context windows beyond hundreds of thousands of tokens~\cite{Chen2023ExtendingCW, ding2024longropeextendingllmcontext, jin2024llmmaybelonglmselfextend}, enabling multi-document and long-context retrieval capabilities such as summarization and question answering. This is handy for technical document understanding, dialogue summarization, and other use cases that demand fine-grained reasoning over increasingly long and complex documents~\cite{liu-etal-2024-lost, li2024longcontextllmsstrugglelong}.

Despite these advancements, recent studies on Needle-in-a-Haystack (NIAH) tasks~\cite{kamradt2023needle, Modarressi2025NoLiMaLE} reveal that many state-of-the-art LLMs struggle to retrieve and reason over granular details embedded deep within long context, often well below the models' claimed maximum context length. Performance degradation becomes especially evident when the relevant information shares superficial similarity with the query, exposing major challenges for tasks requiring associative and semantic reasoning over extended inputs. Complex narrative benchmarks like \novelqa~\cite{novelqa} further highlight this limitation as models must reason about interconnected story elements, character arcs and thematic connections distributed across full-length novels.

To address these limitations, we propose Tagging-Augmented Generation (TAG), a lightweight prompting and input augmentation technique that explicitly highlights salient elements in the input context. By injecting structured, LLM-generated semantic annotations (e.g., named entities, topics, discourse roles) directly into the document, our method helps models focus their attention during inference without requiring any architectural modifications or retrieval infrastructure. Compared to traditional retrieval-augmented generation (RAG) pipelines, semantic tagging provides an interpretable, low-latency and infrastructure-free approach to improve LLM performance on long-context reasoning tasks. Furthermore, tags can be cached, so that context that was previously tagged can be retagged without incurring additional compute overhead.  Figure~\ref{fig:tagthenanswer} shows a high-level comparison between traditional long-context QA and our proposed TAG approach.

\noindent\textbf{Motivation -}  
In real-world scenarios, such as technical support and document analysis, LLMs are often required to process previously unseen, unstructured, and extensive documents without the benefit of pre-indexed retrieval systems. As context lengths grow and becomes more complex, traditional attention mechanisms struggle to maintain focus across distant semantic spans, especially when lexical overlap is minimal~\cite{muennighoff2022sgptgptsentenceembeddings, Modarressi2025NoLiMaLE}. Our semantic tagging approach addresses this gap by embedding explicit structural cues within the input, thereby guiding model attention, preserving information retention and enhancing reasoning quality in long and complex context settings.


\noindent\textbf{Contributions -}
\textbf{(i) Semantic Tagging and Context Enhancement Framework}: We introduce TAG, a flexible framework that employs different tagging mechanisms to produce semantic tags while being agnostic to any named entity recognition (NER) method. TAG supports both traditional and agentic approaches to generate tags without model retraining. To the best of our knowledge, this is the first framework that guides LLMs to perform better in long-context and complex reasoning tasks by semantically enriching the input. \textbf{(ii) Benchmark Extension}: We develop augmented versions of two popular long-context benchmarks, \nolima~\cite{Modarressi2025NoLiMaLE} and \novelqa~\cite{novelqa} to obtain \nolima+ and \novelqa+, respectively. These datasets are purpose-built for evaluating LLMs' ability to utilize structural cues in long contexts. \textbf{(iii) Empirical Validation}: We present a novel study of empirical results in two different environments -- varying context length and complexity settings, using the \nolima+ and \novelqa+ benchmarks. Through extensive experiments using two Anthropic Claude models, we demonstrate that semantic tagging improves question-answering accuracy by over 17\% for 32K token contexts, and 2.9\% in complex reasoning question-answering for multi-hop queries requiring knowledge across a wide span of text, as compared to baselines without tagging.

\section{Background} \label{sec:background}
\noindent \textbf{Needle-in-a-Haystack Evaluation} - The Needle-in-a-Haystack (NIAH)~\citep{kamradt2023needle} paradigm evaluates long-context understanding by embedding specific information within extensive irrelevant content. Traditional NIAH benchmarks rely heavily on lexical overlap between queries and relevant passages, which may not reflect true semantic reasoning capabilities. \nolima~\citep{Modarressi2025NoLiMaLE} addresses this limitation by introducing a benchmark with minimal lexical cues, requiring latent semantic reasoning across long contexts. The benchmark places `needles' containing key information within `haystacks' of random text patches, where questions probe the subject's identity without sharing keywords with the needle. This forces models to rely on semantic associations rather than pattern matching, exposing fundamental limitations in current attention mechanisms even for models with extended context windows.

\noindent \textbf{Complex Narrative Comprehension} - While NIAH benchmarks test information retrieval, real-world applications often require understanding complex, interconnected narratives with multiple characters, plotlines and thematic elements distributed across extensive texts. \novelqa~\citep{novelqa} showcases this through a comprehensive benchmark of 2,305 manually annotated questions spanning 89 full-length novels. This tests models' ability to maintain coherent understanding across extended narratives, requiring multi-hop reasoning about character relationships, plot development and thematic connections. Questions are systematically categorized by complexity (single-hop, multi-hop, detail), enabling granular assessment of long-context comprehension capabilities.

\noindent \textbf{Relevance to our approach -} Our work builds on the above insights by introducing semantic tagging, a prompting strategy that explicitly highlights salient information, helping models better navigate intricate contexts without architectural changes.


\section{Proposed Approach}

\begin{figure}[t]
    \centering
    \includegraphics[width=\linewidth]{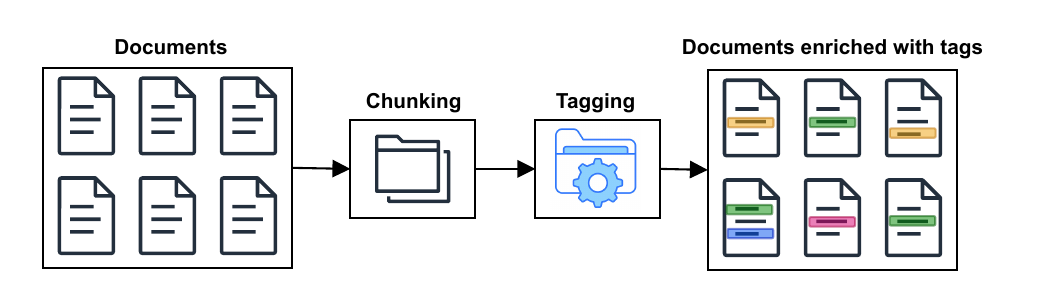}
    \caption{Overview of the proposed TAG framework. Input text is first segmented into de-duplicated chunks using configurable strategies (e.g., sentence-level, paragraph-level, or semantic chunking). These chunks are processed through tagging modules, producing XML-style semantic annotations that preserve document structure while providing explicit attention guidance for downstream tasks. Finally, we have the enriched documents with tags embedded within the text.}
    \label{fig:framework}
\end{figure}

\textbf{TAG Framework} - We propose Tagging-Augmented Generation (TAG), a simple and lightweight prompting framework that mitigates long-context and complex-context performance degradation in LLMs through explicit semantic mark-up. TAG addresses the fundamental challenge of attention degradation in extended contexts by systematically injecting structured semantic annotations into input documents.

The TAG framework operates through a two-stage process. As illustrated in Figure~\ref{fig:framework}, our pipeline starts by segmenting the input text into a list of multi-sentence chunks, using a configurable chunking strategy and granularity. While the simplest strategy involves sentence- or paragraph-level chunking, more advanced techniques such as semantic chunking can also be applied. The resulting chunks are de-duplicated and passed into the respective tagging modules. The output is a list of annotated chunks with semantic tags, which can be used for downstream evaluation or to enhance generation tasks. Semantic elements are annotated using XML-style tags (e.g., \verb|<sem_category>content</sem_category>|) that preserve document structure while providing explicit attention guidance during inference\footnote{The idea behind using XML-based tags to segment context has also been backed by a concurrent and independent work by Anthropic on \href{https://www.anthropic.com/engineering/effective-context-engineering-for-ai-agents}{context engineering}.}.

TAG is method-agnostic, supporting various tagging mechanisms including LLM-based approaches, traditional NLP tools and agentic systems. Unlike traditional retrieval-augmented generation methods that require external knowledge bases~\citep{chen2025enrichindexusingllmsenrich, shen2025rightwayassessingdocument}, TAG operates entirely within the input context, making it infrastructure-free and computationally efficient. The framework particularly excels in scenarios where relevant information exhibits minimal lexical overlap with queries, enabling stronger semantic reasoning across extended contexts without architectural modifications.

\noindent \textbf{Semantic Tagging Methods} - We apply semantic tagging to long-form documents using two approaches: with large language models (LLMs) and using traditional named entity recognition algorithms. Both methods operate over a shared preprocessing pipeline as depicted in Figure ~\ref{fig:framework}.  

\noindent \textbf{(i) LLM-based Tagging using privileged information} - We explore two complementary strategies for semantic tagging using LLMs: Information Extraction (IE)-based tagging and classification-based tagging. Both approaches rely on carefully designed prompts that include persona setup, task instructions, output formatting constraints, and a predefined list of semantic tag categories with their definitions and examples. 

In the IE-based approach, the LLM is prompted  to extract entities from the input text and assign each to one of the defined semantic categories. Each entity is subsequently annotated in the input using a consistent text-based format, such as \texttt{<Person>Marie Curie</Person>}. These tags are inserted directly into the input and retained during inference, enabling the model to more effectively attend to semantically relevant spans when performing tasks such as question answering. A key advantage of this approach is that it eliminates the need for structured output formatting or post-processing such as output parsing. However, it may compromise input fidelity, as the model can hallucinate tags or alter the original text.

In the classification-based approach, we treat each semantic tag category as a separate class, and the prompt instructs the model to identify and output all tags that match the input text. The main advantage of this approach, compared to the IE prompting strategy, is the ability to support few-shot prompting, faster speed (owing to reduced maximum output token generation parameter), and no risk of novel tags or input text changes. The primary trade-offs are the need to enforce a specific output format within the prompt and to implement post-processing logic to extract the identified tags from the model’s response.

To combine the strengths of both methods, we merge their outputs into a unified set of tagged chunks in this work. This hybrid strategy improves overall tag coverage and robustness, while mitigating the limitations of either method in isolation. The system prompts for both IE- and classification-based approaches are shown in Appendix~\ref{appx:sem_tag_prompts}.

\noindent \textbf{(ii) Traditional NER Tagging} - An easy to implement alternative to LLM-based tagging is using Named Entity Recognition (NER) algorithms to extract entities and subsequently tag chunked text. In this work, we use spaCy~\citep{honnibal2020spacy}, an industrial-strength natural language processing library, to perform NER on segmented text data. This model identifies 18 standard entity types, including persons, organizations, and geopolitical entities, etc (see Figure~\ref{apx:entity_mappings}). The implementation employs a nested tagging approach, where text segments containing multiple entity types are wrapped with corresponding XML-style tags, preserving the hierarchical nature of entity relationships. The spaCy tagging approach ensures comprehensive entity coverage while maintaining the contextual integrity of the source materials.

\noindent \textbf{Semantically enhancing documents and benchmarks using TAG} - 
\label{sec:tag_bench}
As discussed in Section~\ref{sec:background}, NIAH and complex narrative benchmarks provide a crucial test bed for revealing the limitations of modern LLMs.  While such limits have been discovered and documented, it is helpful to enhance the test beds with universally applicable, practical alterations that can make the benchmark tasks easier, helping some LLMs and LLM-based agentic solutions gain performance. Towards this end, we demonstrate the applicability of TAG by proposing \nolima+ and \novelqa+, the tag-enriched versions of \nolima~\cite{Modarressi2025NoLiMaLE} and \novelqa~\cite{novelqa} benchmarks, respectively. We note that TAG is not limited to these benchmarks and can be used to semantically improve any type of document using tags.


\nolima+ augments \nolima's test pairs by marking up haystack chunks that match candidate semantic tag categories. Each semantic tag is associated with definitions and examples, and chunks can receive multiple markups when appropriate. As mentioned in the original paper ~\cite{Modarressi2025NoLiMaLE}, the needles are defined using a list of keywords. For our paper, we use these keywords to define our list of privileged tags. We refer to Figure~\ref{fig:privileged_tag_categories} for a complete list of tags and how they are defined in the prompts.

\novelqa+ enhances the \novelqa{} benchmark by applying semantic tagging to the novel texts used for question answering. Unlike \nolima+, which focuses on entity-specific tags relevant to needle-in-haystack retrieval, \novelqa+ employs broader narrative-oriented semantic categories to support complex reasoning across interconnected story elements. We segment each novel into manageable chunks and apply our TAG framework. \novelqa+ enables evaluation of how semantic tagging affects different types of reasoning complexity, from single-hop character identification to multi-hop plot analysis, providing insights into TAG's effectiveness across varying narrative comprehension challenges. In lieu of privileged information, we tag this benchmark using the spaCy entities. The complete list of tags is present in Figure~\ref{fig:spacy_tag_categories}.

 \begin{table*}[t]
  \small
  \centering
  \begin{tabular}{lcccccccc}
    \toprule
   \multirow{2}{*}{Models} & \multirow{2}{*}{\begin{minipage}{0.5in}Tagged Context\end{minipage}} & \multirow{2}{*}{Tagger} & \multirow{2}{*}{\begin{minipage}{0.7in}Tag definition in prompt\end{minipage}} & \multicolumn{4}{c}{Context Length (CL)} & \multirow{2}{*}{\begin{minipage}{0.5in}Extremum drop rate\end{minipage}} \\
    \cmidrule(lr){5-8}
     &  && & 250 & 500 & 16K & 32K & \\
    \midrule
   \multirow{4}{*}{Claude 3.5 Sonnet} & No & - & No & 81.19 & 86.19 & 45.96 & 32.67 & \textcolor{Gray}{59.77}     \\
      & No & - & Yes & \textbf{91.34}   & \textbf{90.58} & \textbf{50.25} & \underline{36.45} & \textcolor{Red}{60.1} \\
         & Yes & spaCy & Yes & \underline{88.77} & \underline{88.36} & 47.65 & 34.63 & \textcolor{Red}{60.99}\\
         & Yes & Privileged & Yes & 87.53 & 85.35 & \underline{48.4} & \textbf{38.5} & \textcolor{OliveGreen}{56.0}\\
         \midrule
   \multirow{4}{*}{Claude 3.7 Sonnet} & No & - & No & 94.66 & 93.48 & 55.76 & 45.56 & \textcolor{Gray}{51.87} \\
       & No & - & Yes & \textbf{97.12} & \textbf{95.7} & 59.22 & \underline{49.93} & \textcolor{OliveGreen}{48.59} \\
       & Yes & spaCy & Yes & 95.11 & 94.52 & \underline{60.39} & 47.88 & \textcolor{OliveGreen}{49.66} \\
       & Yes & Privileged & Yes & \underline{95.21} & \underline{94.77} & \textbf{60.78} & \textbf{52.39} & \textcolor{OliveGreen}{44.98} \\
    \bottomrule
  \end{tabular}
    \caption{Evaluation results on the \nolima+ benchmark. We show the accuracy scores of two Anthropic models, without (baseline) and with assistance via tags, across both the context, and the system prompt. The best scores for a configuration are shown in \textbf{bold}, while the second best are \underline{underlined}. We also show the drop rates of each model from the 250 to 32K context lengths. Baseline results are shown in \textcolor{Gray}{gray}, improvements over baseline are highlighted in \textcolor{OliveGreen}{green} and performance below baseline is shown in \textcolor{Red}{red}. \label{table:sampletable}}
\end{table*}

\section{Experiments}
To comprehensively evaluate the effectiveness of TAG across different long-context scenarios, we conduct two sets of experiments -- (i) long-context needle-in-haystack evaluation using the proposed \nolima+ benchmark (Section~\ref{sec:tag_bench} and Section~\ref{appex:nolima}), which tests semantic retrieval capabilities in the absence of lexical cues and (ii) complex narrative comprehension evaluation using the proposed \novelqa+ benchmark, which assesses reasoning over interconnected story elements across full-length novels (see Section~\ref{sec:tag_bench} and Section~\ref{appex:novelqa}). As many of the novels exceeded the context window limit of Claude models, we had to truncate the books, and filter out questions that were from the omitted section. This left us with 1,035 questions ranging over different levels of complexity (single-hop, multi-hop, and detailed reasoning).

Through our experiments, we investigate the following research questions: \textbf{RQ1}: Can semantic tagging improve LLM performance on long-context retrieval tasks without relying on greedy search-based methods or architectural modifications? \textbf{RQ2}: How do different tagging methodologies, such as providing tag definitions alone versus augmenting the context with explicit tags, impact long-context performance and degradation rates? \textbf{RQ3}: Does tagging help both non-reasoning and reasoning model performance?

\noindent \textbf{Model Setup} - For generating responses to questions based on untagged and tagged data, we consider two recent models from the Anthropic Claude families -- Claude Sonnet 3.5 v2 \cite{claude35} and Claude Sonnet 3.7 \cite{claude37}. Each of these models support a long context window of 200K tokens, thereby making them suitable for NIAH tasks such as \nolima{} and \novelqa. Additionally, these models rank among the best in terms of performance on long-query (i.e. >500 tokens) results in the LLMArena leaderboard \cite{chiang2024chatbotarenaopenplatform}. We invoke both these models through Amazon Bedrock, integrating them directly into our codebase for automated inference over long documents. For Claude 3.5 Sonnet, we keep \texttt{temperature = 0} to make it as deterministic as possible in its prediction, while for Claude 3.7 Sonnet, we enable the \texttt{thinking} feature, which supersedes the choice of temperature.

\noindent \textbf{Evaluation Metrics} - For a fair comparison with the \nolima{} benchmark, we adopt the same evaluation metrics. More specifically, we give an LLM a score of 1 as long as the generated response \textit{contains} the golden answer, else 0. Furthermore, we conduct our experiments using context lengths of 250, 500, 16K and 32K tokens respectively\footnote{Our approach is not limited by the context length, but we want to emphasize the utility of tagging really long contexts like 16K and 32K tokens, by benchmarking them against shorter contexts of token sizes 250 and 500.}. 

The evaluation methodology for \novelqa{} follows a straightforward multiple-choice assessment framework. We evaluate the model performance using \textit{exact match} accuracy by comparing the LLM's single-character output (A, B, C, or D) against the golden answer provided in the \novelqa{} dataset. 

\noindent \textbf{Experimental settings} - We evaluated two semantic tagging approaches for the long-context understanding tasks:
 \textbf{(i) Tag Definitions only (TD)}: We hypothesize that providing explicit definitions of key semantic elements within the task prompt can enable models to better identify relevant information across increasing context lengths. \textbf{(ii) Tag Definitions with Tagged Context (TD+TC)}: Our second experimental setting involves combining tag definitions in the prompt with guided context navigation using tagged context. By strategically applying semantic tags throughout the context and providing explicit definitions of these tags, we expect models to maintain stronger associative reasoning capabilities even at the longest context lengths. The baseline model in our experiments is the vanilla RAG setting with NO tagged context, and NO tag definitions in prompt. 

\begin{table*}[htpb]
  \small
  \centering
  \begin{tabular}{lccccccc}
    \toprule
   \multirow{2}{*}{Models} & \multirow{2}{*}{\begin{minipage}{0.5in}Tagged Context\end{minipage}} & \multirow{2}{*}{Tagger} & \multirow{2}{*}{\begin{minipage}{0.7in}Tag definition in prompt\end{minipage}} & \multicolumn{3}{c}{Complexity}  \\
    \cmidrule(lr){5-7}
     &  &  & & Single-hop & Multi-hop & Detail   \\
    \midrule
   \multirow{3}{*}{Claude 3.5 Sonnet} & No & - & No & \underline{86.88} & 48.71 & 73.36    \\
      & No & - & Yes &  86.39 &  \textbf{50.12} &  \underline{74.77} \\
         & Yes & spaCy & Yes & \textbf{87.87} & \underline{49.88} & \textbf{78.97} \\
         \midrule
   \multirow{3}{*}{Claude 3.7 Sonnet} & No & - & No & 90.10 & \underline{56.72} & \textbf{84.11*} \\
       & No & - & Yes & \underline{90.84} &  56.23 &  \underline{82.71} \\
       & Yes & spaCy & Yes & \textbf{91.09*} & \textbf{56.97*} & 81.31 \\
    \bottomrule
  \end{tabular}
    \caption{Evaluation results on the \novelqa+ benchmark. We show the accuracy scores of two Anthropic models, without (baseline) and with assistance via tags, across both the context and the system prompt. The best scores are shown in \textbf{bold}, while the second best are \underline{underlined}.  Overall highest accuracy is marked with an asterisk.\label{table:complexitytable}}
\end{table*}


\noindent \textbf{Evaluation Results and Discussion} - Table \ref{table:sampletable} presents the accuracy scores across various context lengths ranging from 250 to 32K tokens. In the baseline condition (i.e., no semantic tagging and no tag definition in prompt), Claude 3.5 Sonnet's performance dropped from 81.19\% at CL250 to 32.67\% at CL32K, representing a 59.77\% extremum drop rate. Similarly, Claude 3.7 Sonnet experienced a 51.87\% extremum drop rate from 94.66\% at CL250 to 45.56\% at CL32K.  

The TD approach improved short-context performance for both models compared to the baseline, with Claude 3.5 Sonnet reaching 91.34\% at CL250 (+10.15 gain) and Claude 3.7 Sonnet reaching 97.12\% (+2.46 gain). When incorporating SpaCy tagging, the models achieved 88.77\% and 95.11\% at CL250 for Claude 3.5 and 3.7 respectively. This demonstrates that presence of clear task definitions in the prompt could significantly enhance models' ability to establish semantic connections (\textbf{RQ2}). However, this approach did not substantially reduce the extremum drop rate, with Claude 3.5 Sonnet still experiencing a 60.1\% decline, SpaCy showing a 60.99\% decline, and Claude 3.7 Sonnet a 48.59\% decline from peak performance to CL32K. 

While the privileged tagging approach (TD + TC) showed slightly lower peak performance compared to TD alone for CL250 (87.53\% vs 91.34\% for Claude 3.5, 95.21\% vs 97.12\% for Claude 3.7), it significantly reduced degradation at CL32K, with Claude Sonnet 3.7 achieving 52.39\% performance accuracy, compared to 49.93\% with TD alone and 45.56\% in the baseline (\textbf{RQ1}). This suggests that tagging becomes increasingly effective as context length extends beyond certain thresholds. Figure~\ref{fig:claude3.7_tag_thinking} shows an illustration of Claude 3.7's ability to reason about the tags present in the context to accurately extract the correct answer.


Table \ref{table:complexitytable} presents the evaluation results for the \novelqa{}     dataset. The experimental analysis reveals distinct patterns across different model configurations and complexity categories. In the baseline configuration (Section~\ref{appex:user_prompts}), Claude 3.7 Sonnet demonstrates superior performance compared to 3.5 Sonnet v2 across all complexity categories (90.10\% vs 86.88\% for single-hop, 56.72\% vs 48.17\% for multi-hop, and 84.11\% vs 73.36\% for detailed questions). With tag definitions only (TD) (Figure~\ref{fig:untagged_tag_def_True}), both models show modest improvements in certain categories, with Claude 3.7 achieving 90.84\% for single-hop questions and Claude 3.5 Sonnet showing improved performance in multi-hop scenarios (50.12\% vs baseline 48.17\%). The integration of TD+TC with spaCy (Figure~\ref{fig:tagged_tag_def_True}) yields the most promising results, particularly for Claude 3.7 Sonnet, which achieves the highest accuracy across single-hop (91.09\%) and multi-hop (56.97\%) scenarios. Notably, Claude 3.5 Sonnet v2 shows substantial improvement in detailed questions under this configuration, reaching 78.97\% accuracy compared to its baseline of 73.36\%. These results suggest that the combination of TD+TC can enhance model performance, particularly for more complex reasoning tasks. Considering the above observations in both experiments, semantic tagging using TAG improves the performance of LLMs in both non-reasoning and reasoning settings (\textbf{RQ3}).


\section{Conclusion}

We proposed TAG, a lightweight prompting and input augmentation technique that explicitly highlights salient elements in long and complex contexts through structured semantic annotations. By injecting XML-style tags directly into input documents, our method helps models focus their attention during inference without requiring architectural modifications or retrieval infrastructure. TAG is method-agnostic, supporting various tagging mechanisms including LLM-based approaches, traditional NLP tools, and agentic systems, making it infrastructure-free and computationally efficient.

Through extensive experiments on \nolima+ and \novelqa+ benchmarks, we demonstrate that semantic tagging improves question-answering accuracy by over 17\% for 32K token contexts and 2.9\% in complex reasoning tasks, while significantly reducing performance degradation rates.  Our results establish TAG as a practical and scalable approach for strengthening LLM performance in extended inference scenarios without the computational overhead of traditional retrieval systems.

\section*{Limitations} 

While TAG shows promising results, several areas warrant further investigation. Our evaluation focuses on question-answering tasks within synthetic (\nolima+) and literary (\novelqa+) domains; broader validation across diverse tasks and technical domains would strengthen our findings. The framework's effectiveness depends on appropriate semantic category design, which may present challenges in specialized domains or low-resource languages.

TAG introduces modest computational overhead during preprocessing, though this remains lighter than traditional retrieval systems. While our method enhances attention to relevant content, it operates within existing context windows and cannot address fundamental memory limitations of current LLMs. We note that our experiments primarily utilized spaCy-based and LLM-based tagging; comprehensive evaluation of agentic tagging approaches on-the-fly at inference presents an important direction for future work.


\bibliography{custom}

\clearpage
\appendix

\section{Appendix}


\subsection{Background and Related Work}

\subsubsection{Long-Context Challenges in LLMs}
Despite the advancements enabling LLMs to process extended sequences, empirical evidence indicates persistent challenges in effectively utilizing long contexts. Several factors contribute to LLM failures in long-context settings, including attention degradation, lost-in-the-middle and literal match dependence. We briefly describe these factors below.

\textbf{Attention Degradation.} The softmax-based self-attention mechanism in Transformer architectures struggle to maintain focus as context length increases. This causes diluted attention weights, particularly for the tokens in the middle of the input sequences \cite{Liu2023LostIT}.

\textbf{Lost-in-the-Middle.} This phenomenon describes a systematic attention bias where language models struggle to accurately retrieve and utilize information positioned in the middle portions of long contexts. Recent empirical studies have demonstrated that tokens located centrally within extensive sequences receive disproportionately less attention compared to those at the beginning and end~\cite{Liu2023LostIT}. This positional bias leads to a significantly reduced performance for content placed in middle segments. The effect becomes increasingly evident as context length expands~\cite{zhang2024middlelanguagemodelsuse}. 

\textbf{Literal Match Dependence.} LLMs often demonstrate dependence on literal lexical overlap, succeeding primary when queries and relevant passages share explicit term matches, which limits their ability to generalize to semantically similar, but lexically distinct contexts \cite{Modarressi2025NoLiMaLE, sun2021longrangelanguagemodelsactually}. Particularly, the \nolima{} benchmark~\cite{Modarressi2025NoLiMaLE} illustrates these issues by removing literal cues and requiring associative reasoning. Even top-tier models like GPT-4o drop from 99.3\% accuracy at 1K tokens to 69.7\% at 32K tokens on \nolima.

\subsubsection{Input Tagging and Modular Prompt Conditioning}
Our work is closely related to that of Tag-LLM~\cite{shen2024tagllmrepurposinggeneralpurposellms}, which introduces a framework for repurposing general-purpose language models for specialized domains via learnable input tags. These tags are continuous embeddings prepended to the model’s input sequence to encode domain- or task-specific information.  Tag-LLM learns two types of tags, i.e. domain tags and function tags, through a hierarchical protocol that leverages both unlabeled and labeled data. Notably, Tag-LLM retains these embedding-level tags during inference, conditioning the model throughout the input sequence. In contrast, our semantic tagging approach operates purely at the textual level, using LLM-generated markers to highlight salient content. While both methods retain tags during inference to guide model behavior, Tag-LLM relies on fine-tuned internal conditioning, whereas our approach offers a lightweight, interpretable alternative that requires no model modifications or additional training.

\subsection{Datasets}
\subsubsection{\nolima}\label{appex:nolima}
Our experiments utilize the \nolima{} (No Literal Matching) benchmark~\citep{Modarressi2025NoLiMaLE}, a recently developed evaluation framework specifically designed to assess semantic understanding capabilities of LLMs in long-context scenarios. Unlike traditional long-context benchmarks that rely on lexical overlap between questions and relevant content, \nolima{} deliberately minimizes literal matches, requiring models to make latent semantic connections to succeed. Each evaluation instance consists of a `needle' (i.e., a sentence containing key information about a character and a concept), a `question' that refers to the needle using semantically related but lexically distinct keywords and a `haystack' of irrelevant text. For example, a needle might state `Yuki lives next to the Semper Opera House' while the question asks `Which character has been to Dresden?', requiring the model to recognize the semantic association between `Semper Opera House' and `Dresden' without surface-level matching. 

This absence of literal matching creates a particularly challenging test case for our research, as it forces models to rely on their deeper semantic reasoning rather than pattern matching abilities. The benchmark comprises 58 needle-question pairs with varying complexity levels, including both one-hop and two-hop reasoning paths. To evaluate the performance across different context lengths, each needle is placed at 26 fixed positions throughout haystacks exceeding 60K tokens, constructed from concatenated snippets of open-licensed books. The benchmark's design specifically addresses limitations in existing evaluation frameworks that inadvertently allow models to succeed through surface-level cues rather than true semantic understanding, making it ideal for evaluating our semantic tagging approach.

\subsubsection{\novelqa}\label{appex:novelqa}
\novelqa{} is a benchmark dataset for evaluating long-text language models' ability to understand and answer question about novels. Distinguished by its manually annotated questions crafted by domain experts, the dataset comprises 2,305 questions spanning 89 novels, with a robust taxonomy classifying queries by both complexity (multi-hop, single-hop, and detail) and aspect (times, meaning, span, setting, relation, character, and plot). The benchmark's architecture facilitates comprehensive evaluation through both generative and multiple-choice paradigms, enabling granular assessment of models' capabilities in long-range comprehension and reasoning. The \novelqa{} has the rigorous methodology in dataset construction and evaluation. The questions are deliberately designed to test various of text comprehension while maintaining high-quality control through expert annotation and validation. This framework not only provides quantitative metrics for model performance but also offers insights into specific challenges in long-context understanding, such as evidence position effects and information recall capabilities. The benchmark thus serves as a crucial tool for advancing our understanding of language models' limitations and capabilities in processing extended narratives, contributing significantly to the development of more robust long-context language models.

\subsection{\nolima+ Dataset Creation Pipeline}
\label{apx:nolimaplus_pipeline}

The \nolima+ dataset creation process involves a systematic multi-stage pipeline designed to enhance the original \nolima{} benchmark with semantic tagging capabilities. The enhanced dataset enables more targeted evaluation of retrieval-augmented generation systems by providing structured semantic annotations that guide attention to relevant content sections. We start by segmenting each haystack into a list of multi-sentence chunks, using a predefined chunking strategy and granularity, which is defined as the `maximum chunk size'.The simplest chunking strategy is so-called sentence or paragraph chunking, but it could easily be any other chunking technique, e.g. semantic chunking. 

The list of chunks is then de-duplicated, and each chunk is passed to two complementary LLM-based tagging sub-pipelines. The key difference between these is their prompting strategy. In addition to the chunk text, each of the two sub-pipelines' prompts also includes persona, task and formatting instructions, and a list of the candidate semantic tag categories, their definitions, and examples. 

After output parsing and post-processing steps, the outputs of the two sub-pipelines are merged into a single tagged chunk. All tagged chunks are concatenated to derive the tagged haystacks, which can be then be used to benchmark Tagging-Augmented Generation of arbitrary LLMs or AI Agents.

\tcbset{
    examplebox/.style={
        colback=blue!5,
        colframe=blue!75!black,
        coltitle=white,
        fonttitle=\bfseries,
        boxrule=0.8pt,
        arc=3pt,
        left=4pt,
        right=4pt,
        top=4pt,
        bottom=4pt
    },
    promptbox/.style={
        colback=gray!5,
        colframe=gray!75!black,
        coltitle=white,
        fonttitle=\bfseries,
        boxrule=0.6pt,
        arc=2pt,
        left=3pt,
        right=3pt,
        top=3pt,
        bottom=3pt,
        toptitle=2pt,
        bottomtitle=2pt,
        before skip=8pt plus 2pt minus 2pt,  
        after skip=8pt plus 2pt minus 2pt    
    },
    codebox/.style={
    colback=gray!5,
    colframe=gray!75!black,
        coltitle=white,
        fonttitle=\bfseries,
        boxrule=0.8pt,
        arc=3pt,
        left=4pt,
        right=4pt,
        top=4pt,
        bottom=4pt
    }
}

\tcbset{
    examplebox/.style={
        colback=gray!5,
        colframe=gray!75!black,
        coltitle=white,
        fonttitle=\bfseries,
        boxrule=0.6pt,
        arc=2pt,
        left=3pt,
        right=3pt,
        top=3pt,
        bottom=3pt,
        toptitle=2pt,
        bottomtitle=2pt,
        before skip=6pt plus 2pt,
        after skip=6pt plus 2pt
    },
 promptbox/.style={
        colback=gray!5,
        colframe=gray!75!black,
        coltitle=white,
        fonttitle=\bfseries,
        boxrule=0.6pt,
        arc=2pt,
        left=4pt,
        right=4pt,
        top=4pt,
        bottom=4pt,
        toptitle=3pt,
        bottomtitle=3pt,
        before skip=10pt plus 2pt minus 1pt,
        after skip=8pt plus 2pt minus 1pt
    },
    codebox/.style={
        colback=green!5,
        colframe=green!75!black,
        coltitle=white,
        fonttitle=\bfseries,
        boxrule=0.6pt,
        arc=2pt,
        left=3pt,
        right=3pt,
        top=3pt,
        bottom=3pt,
        toptitle=2pt,
        bottomtitle=2pt,
        before skip=6pt plus 2pt,
        after skip=6pt plus 2pt
    }
}
\raggedbottom

\subsection{Prompts Used}\label{appex:prompts}

\subsubsection{User prompts}\label{appex:user_prompts}
\begin{figure}[ht]
    \centering
    \includegraphics[width=\linewidth]{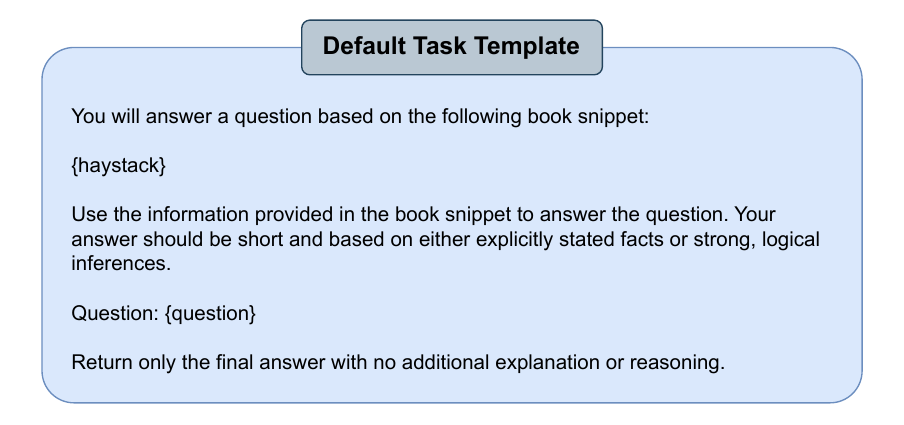}
    \caption{User prompt for NoLima+ dataset.}
    \label{fig:nolima_system_prompt}
\end{figure}

\begin{figure}[ht]
    \centering
    \includegraphics[width=\linewidth]{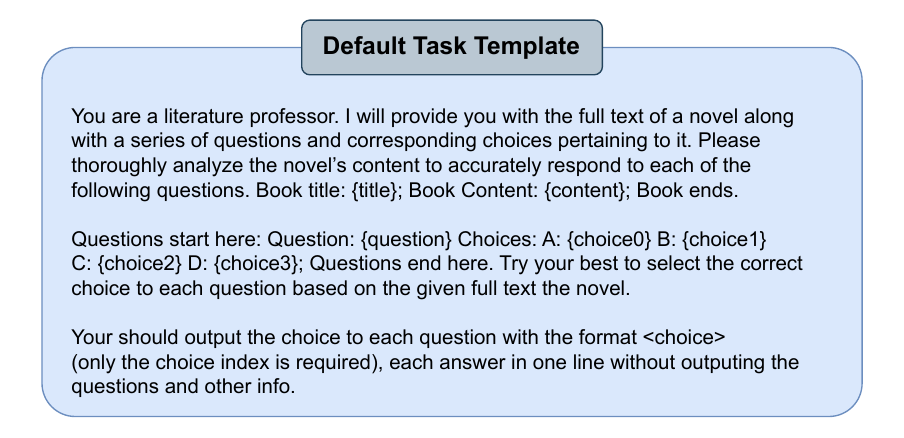}
    \caption{User prompt for NovelQA+ dataset.}
    \label{fig:novelqa_system_prompt}
\end{figure}

\subsubsection{System prompts for TAG framework}
\begin{figure}[ht]
    \centering
    \includegraphics[width=\linewidth]{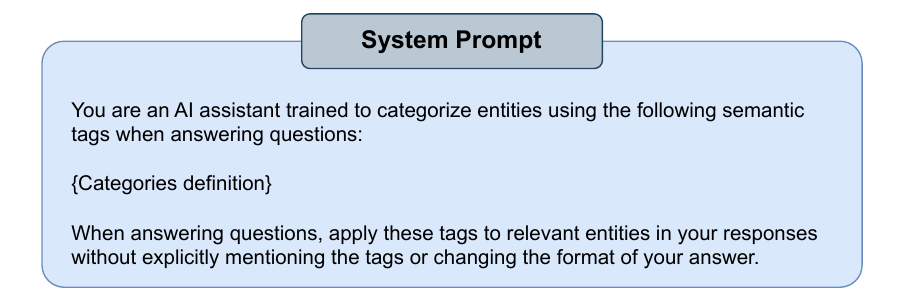}
    \caption{System prompt for untagged context with tag definition in prompt.}
    \label{fig:untagged_tag_def_True}
\end{figure}

\begin{figure}[ht]
    \centering
    \includegraphics[width=\linewidth]{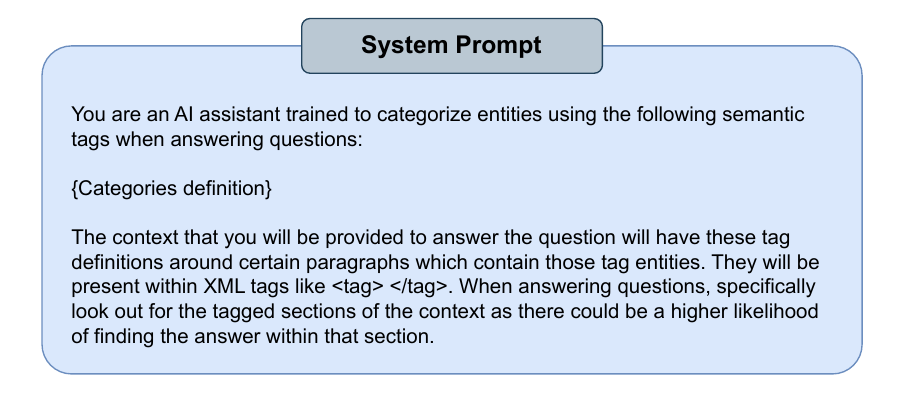}
    \caption{System prompt for tagged context with tag definition in prompt.}
    \label{fig:tagged_tag_def_True}
\end{figure}

\subsubsection{Semantic tagging prompts}\label{appx:sem_tag_prompts}
\begin{figure}[ht]
    \centering
    \includegraphics[width=\linewidth]{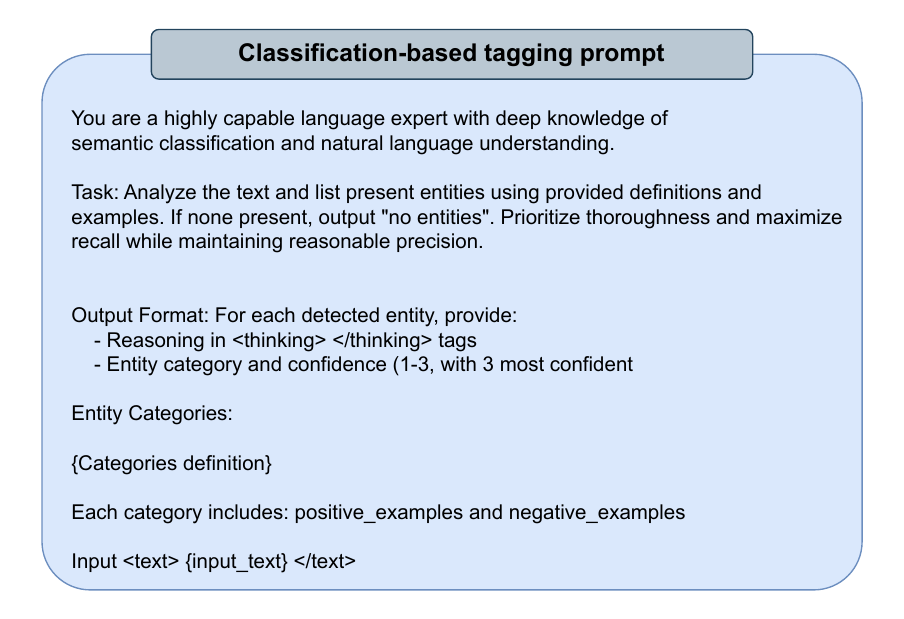}
    \caption{Classification-based tagging prompt.}
    \label{apx:class_tag_prompt}
\end{figure}

\begin{figure}[ht]
\vspace{-0.5in}
    \centering
    \includegraphics[width=\linewidth]{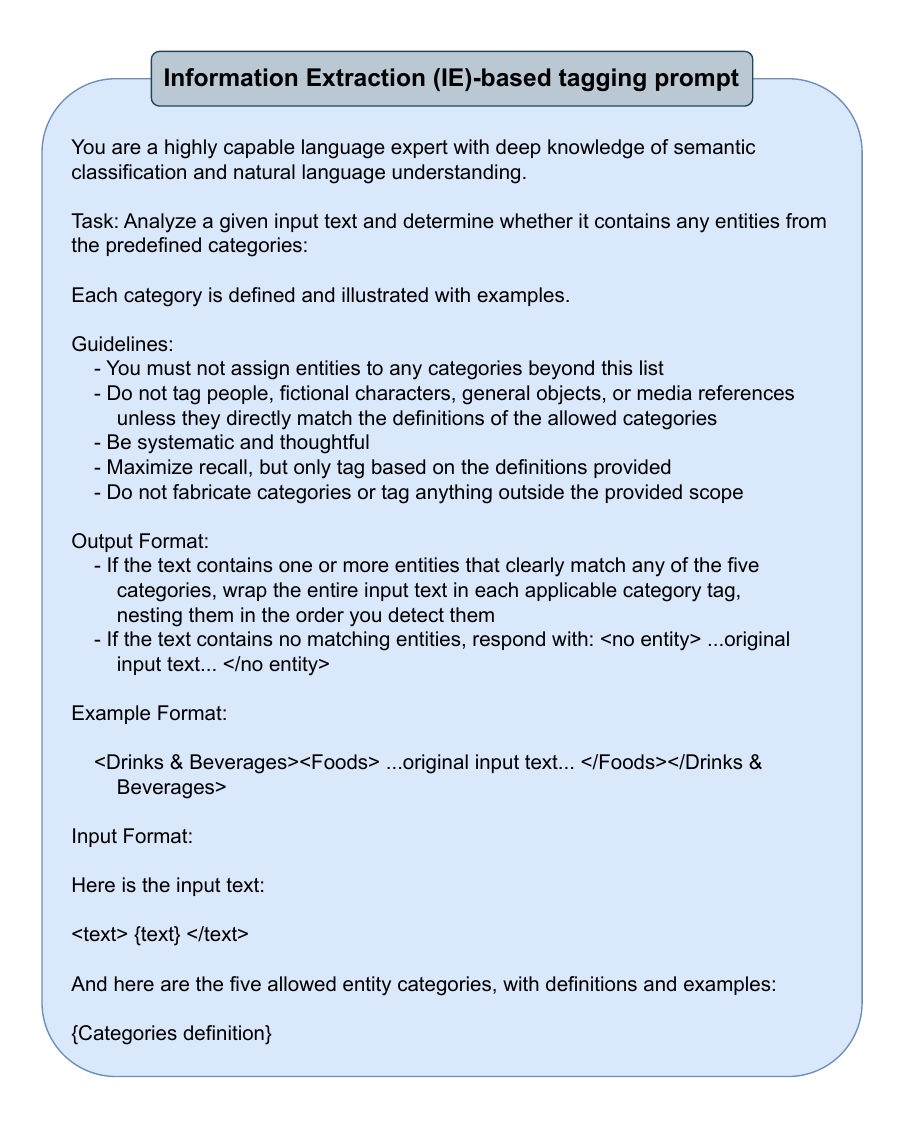}
    \caption{Information Extraction (IE)-based tagging prompt.}
    \label{apx:ie_tag_prompt}
\end{figure}

\subsubsection{Tag category definitions}
\begin{figure}[ht]
    \centering
    \includegraphics[width=\linewidth]{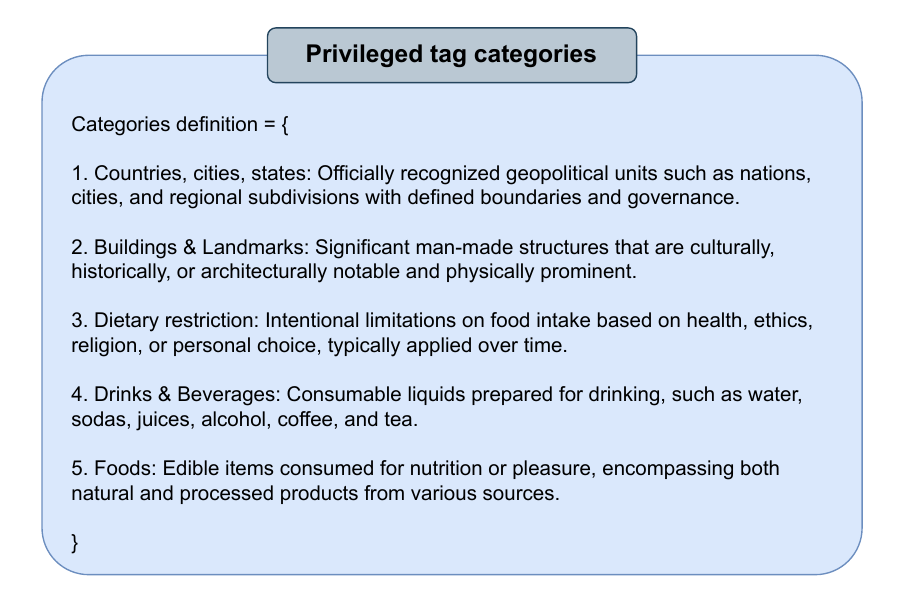}
    \caption{Privileged tag categories.}
    \label{fig:privileged_tag_categories}
\end{figure}
\begin{figure}[t]
    \centering
    \includegraphics[width=\linewidth]{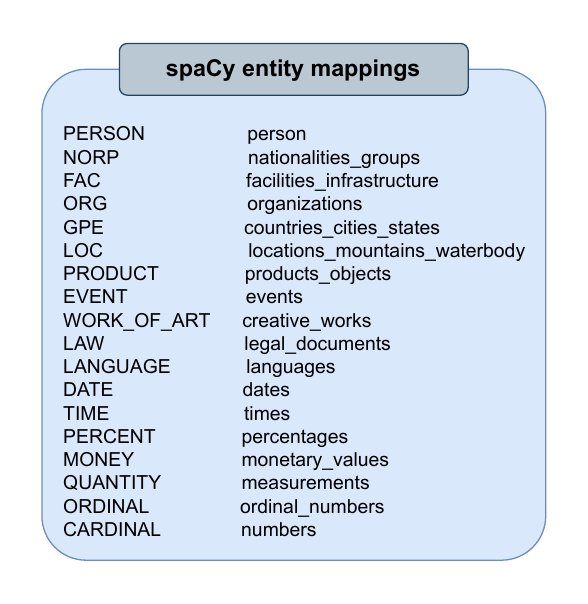}
    \caption{spaCy entity mappings.}
    \label{apx:entity_mappings}
\end{figure}


spaCy provides a list of default named entities which can be mapped to semantic categories. These are shown in Figure~\ref{apx:entity_mappings}.

\begin{figure*}[ht]
    \centering
    \includegraphics[width=0.7\linewidth]{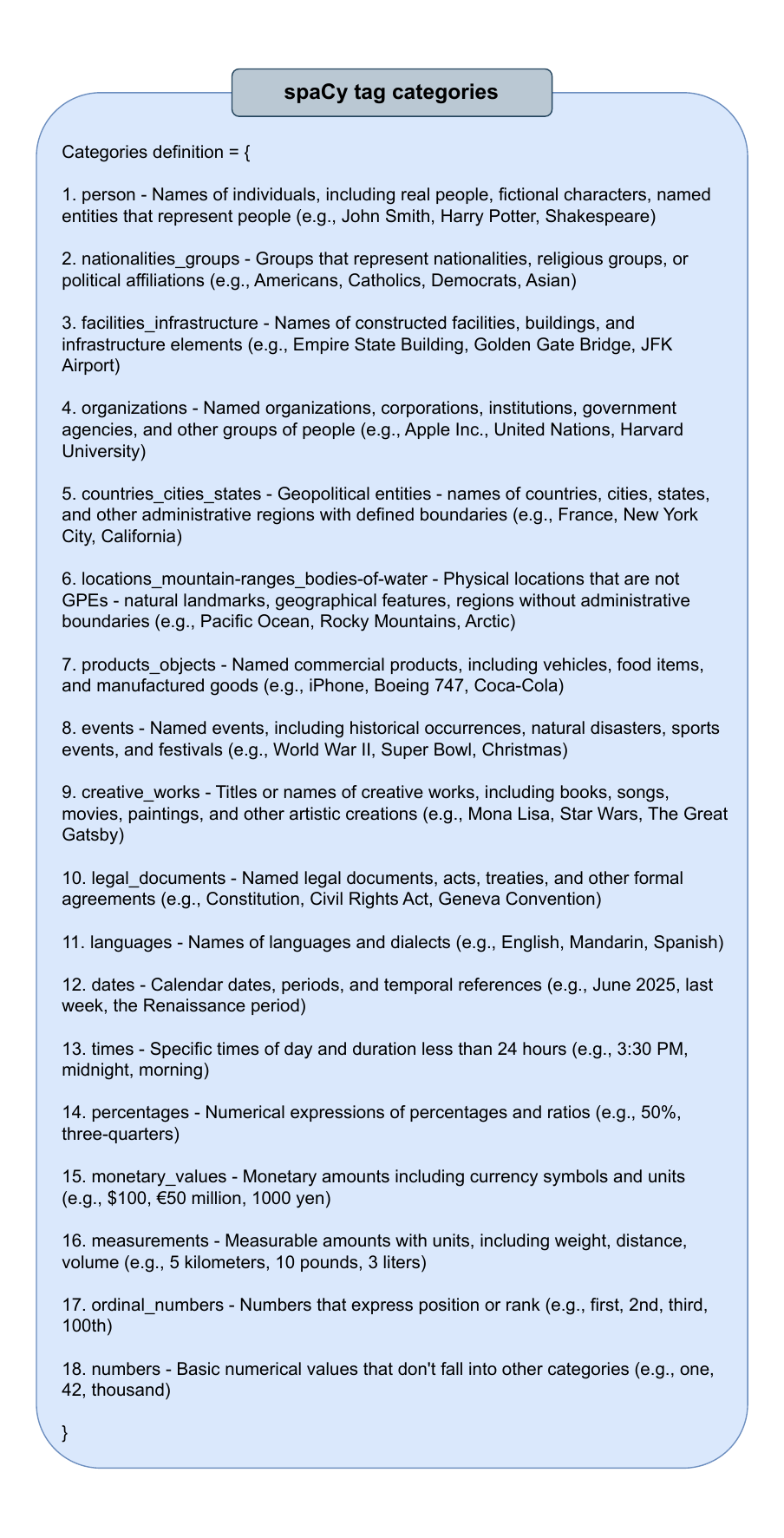}
    \caption{spaCy tag categories.}
    \label{fig:spacy_tag_categories}
\end{figure*}

\subsection{Illustration of TAG helping LLM reasoning}
\begin{figure*}[!t]
  \centering
  \includegraphics[width=\textwidth]{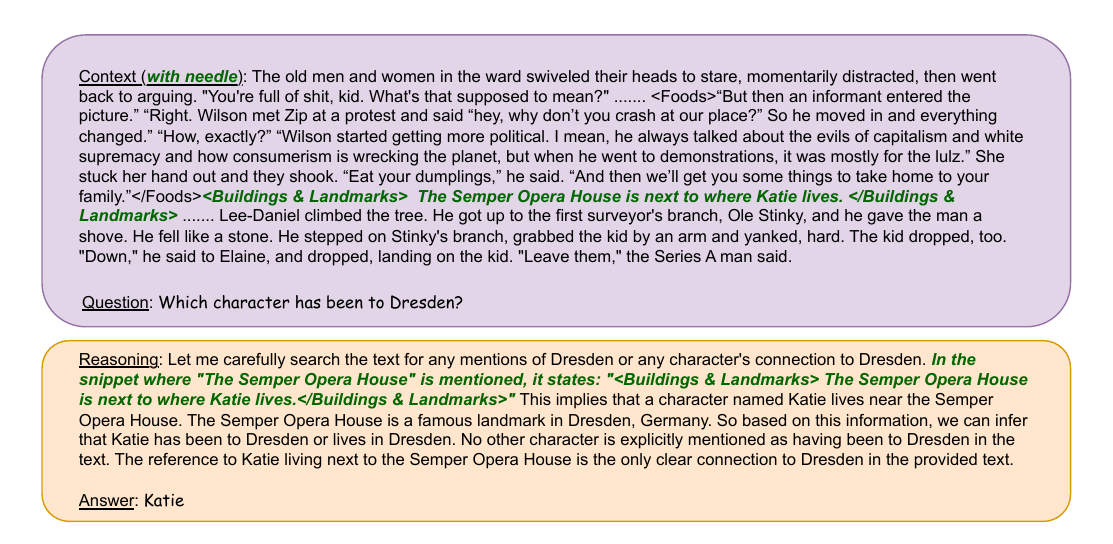}
  \caption{[Best viewed in color] Illustration of the benefits of tagging context in generating correct responses by Claude 3.7 Sonnet. In the above example, a long context with an embedded needle sentence has been tagged by our proposed approach. As seen in the generating reasoning statement, this helps the LLM to attend to the correct part of the context to easily extract the answer to an otherwise difficult question. }
  \label{fig:claude3.7_tag_thinking}
\end{figure*}

\end{document}